\documentclass[conference]{IEEEtran}
\IEEEoverridecommandlockouts
\usepackage{cite}
\usepackage{amsmath,amssymb,amsfonts}
\usepackage{algorithmic}
\usepackage{graphicx}
\usepackage{textcomp}
\usepackage{xcolor}
\usepackage{kotex}
\usepackage{booktabs}
\usepackage{threeparttable}
\usepackage{multirow}
\usepackage[table]{xcolor}
\usepackage{makecell}
\usepackage{array}
\usepackage[ruled,vlined,linesnumbered]{algorithm2e}
\usepackage[colorlinks=true, linkcolor=blue, citecolor=blue]{hyperref}
\usepackage{enumitem}
\def\BibTeX{{\rm B\kern-.05em{\sc i\kern-.025em b}\kern-.08em
    T\kern-.1667em\lower.7ex\hbox{E}\kern-.125emX}}
\begin{document}

\title{NVSim: Novel View Synthesis Simulator for Large Scale Indoor Navigation}

\author{Mingyu Jeong, Eunsung Kim, Sehun Park and Andrew Jaeyong Choi\thanks{All authors are with the School of Computing, Gachon University, 1342 Seongnam-daero, Sujeong-gu, Seongnam 13120, Republic of Korea. \par *Andrew Jaeyong Choi is the corresponding author. Email: \texttt{andrewjchoi@gachon.ac.kr} \par Student emails:\texttt{\{jkg7170, kes2387, sehunpark\}@gachon.ac. kr}}}

\maketitle

\begin{abstract}
We present NVSim, a framework that automatically constructs large-scale, navigable indoor simulators from only common image sequences, overcoming the cost and scalability limitations of traditional 3D scanning. Our approach adapts 3D Gaussian Splatting to address visual artifacts on sparsely-observed floors—a common issue in robotic traversal data. We introduce Floor-Aware Gaussian Splatting to ensure a clean, navigable ground plane, and a novel mesh-free traversability checking algorithm that constructs a topological graph by directly analyzing rendered views. We demonstrate our system's ability to generate valid, large-scale navigation graphs from real-world data. A video demonstration is avilable at \url{https://youtu.be/tTiIQt6nXC8}.
\end{abstract}

\section{Introduction}
The advancement of Vision-and-Language Navigation (VLN) has followed the emergence of realistic indoor simulators. These simulators, reconstructed from precise scans of real spaces and high-quality 3D meshes, provided a foundation for agents to repeatedly learn and evaluate the process of navigating with language instructions \cite{anderson2018vision, chang2017matterport3d}. Therefore, VLN research advanced rapidly in various aspects, including language-vision grounding, long-horizon planning, and generalized policy learning.
However, existing indoor benchmarks and simulators often rely on expensive 3D scanning equipment, manual modeling, and pre-defined sparse viewpoint graphs. This approach results in cost and time constraints for scaling to new locations and imposes a fundamental limitation on the diversity of agent-traversable paths.
Thus, we present \textbf{NVSim}: a framework that automatically constructs large-scale, navigable indoor environments using only common traversal image sequences, without expensive scanning equipment or an explicit mesh generation process. The core idea is to learn a continuous 3D scene representation from an image sequence and, based on this, explore the traversable space to generate a topological graph for navigation.
\noindent Our contributions are as follows:
\begin{itemize}
\item \textbf{Novel View Synthesis Simulator (NVSim):} We propose a new framework that scalably and automatically constructs large-scale indoor environments from only common traversal image sequences.
\item \textbf{Floor-Aware Gaussian Splatting:} To solve the artifact problem that occurs in floor regions during 3D scene representation, we introduce a robust floor segmentation technique and a Floor-aware Loss.
\item \textbf{Mesh-Free Traversability checking:} We propose a method to infer traversability using only rendered views and a zero-shot vision model, without an explicit 3D mesh, and automatically construct a topological navigation graph.
\end{itemize}

\begin{figure}[t]
    \centering
    \includegraphics[width=1.0\linewidth]{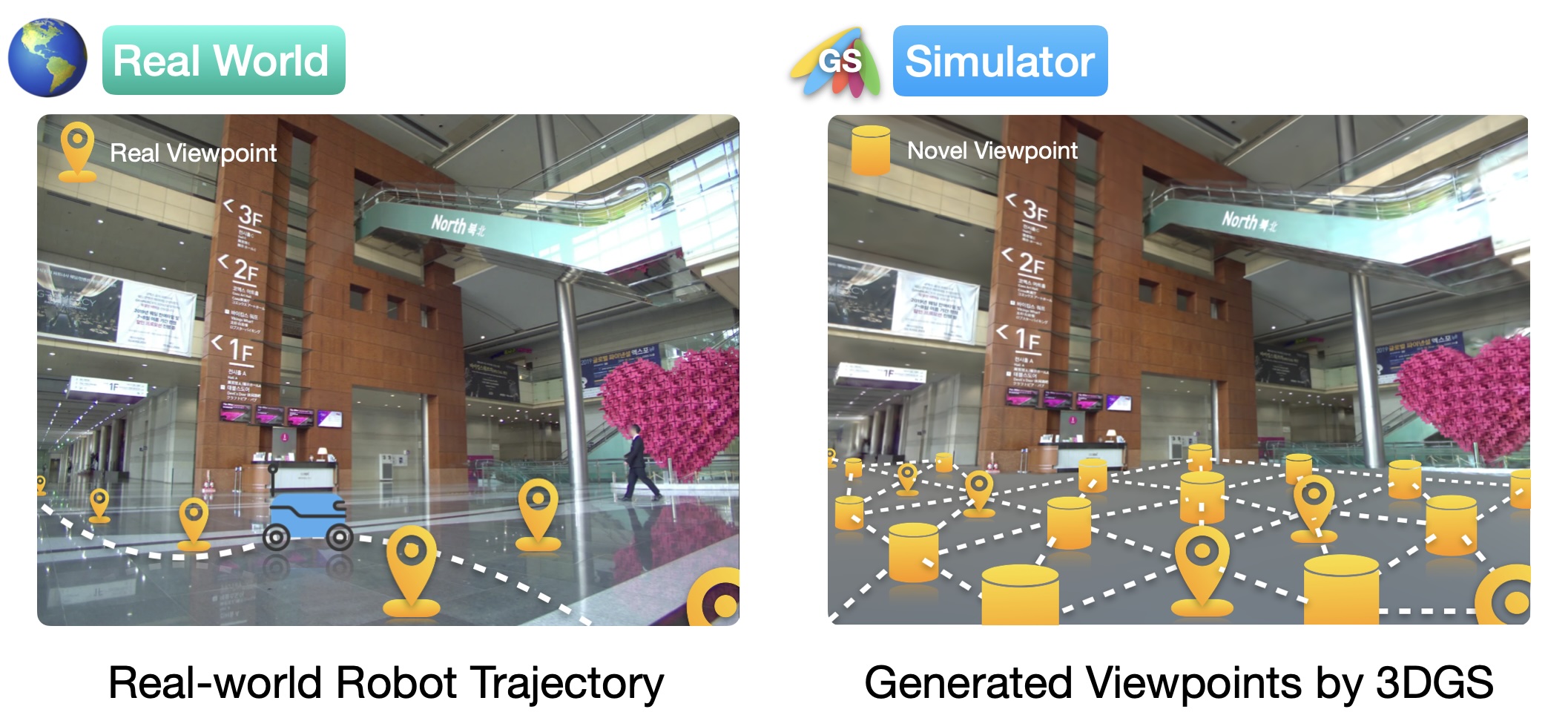}
    \caption{The left panel shows the viewpoints captured along a real-world robot trajectory. The right panel displays the dense graph of traversable viewpoints automatically generated by our method.}
    \label{fig:instr_fig}
\end{figure}
\section{Related Work}

\subsection{Vision-Language Navigation and Simulators}

Visual-Language Navigation (VLN) agents are typically trained and evaluated in simulated environments. Early benchmarks, such as Room-to-Room (R2R) \cite{anderson2018vision}, were constructed from Matterport3D scans and established foundational research in the field. However, these simulators often restrict agent movement to a discrete viewpoint graph and can suffer from capture bias, where views are optimized for visual appeal rather than realistic robotic traversal \cite{anderson2018vision}. While subsequent datasets like RxR \cite{ku2020room} and SOON \cite{zhu2021soon} introduced more complex tasks, they were largely built upon the same static indoor scenes.
To support more realistic interaction, mesh-based simulators like Habitat \cite{savva2019habitat} and iGibson \cite{shen2021igibson} were developed, enabling continuous agent control. This paradigm has been advanced by large-scale, high-fidelity datasets such as HM3D \cite{ramakrishnan2021habitat} and ScanNet++ \cite{yeshwanth2023scannetpp}. Yet, these approaches fundamentally rely on explicit 3D assets, requiring precise mesh and texture data that is typically acquired through expensive 3D scanning. This makes it difficult for researchers to create or customize environments, limiting research to pre-existing datasets.
This scalability challenge is particularly relevant for recent approaches using Large Language Models (LLMs) for zero-shot planning \cite{chen2024mapgpt, zhou2024navgpt, qiao2025open}. These models are trained on large-scale data and have the potential to navigate beyond the residential domains common in existing benchmarks. However, the difficulty in creating new simulation environments limits the ability to evaluate their generalization performance in more varied settings, such as large public spaces. This motivates the need for a low-cost, scalable method to generate new simulators. Our work presents such a method, constructing navigable environments directly from image sequences without requiring explicit mesh data.
\subsection{3D Scene Representation}
Recent advances in 3D scene representation, such as Neural Radiance Fields (NeRF) \cite{mildenhall2021nerf} and 3D Gaussian Splatting (3DGS) \cite{kerbl2023gaussiansplatting}, have enabled new applications in Embodied AI. Prior research has shown direct robot navigation within pre-trained NeRF models \cite{adamkiewicz2022vision} using the density field for collision avoidance, and more recently, real-time motion planning \cite{chen2025splat} and visual goal tracking \cite{zhu2025vr} with 3DGS. Inspired by these advances, we build upon these methods to reconstruct large-scale environments from robot traversal data, with the goal of creating navigation simulators for high-level tasks like Vision-and-Language Navigation (VLN).

However, reconstructing scenes from robot traversal data for navigation simulators faces several challenges. Image sequences often contain transient objects, and the common forward-facing camera configuration leads to sparse observations of the floor. These factors can cause visual artifacts, such as floaters and geometric inconsistencies, in the final reconstruction. If these artifacts occur on the floor, they degrade an agent's perception and hinder navigation. Prior work in 3D scene representation addresses challenges such as transient objects \cite{xu2024splatfacto}, scalability\cite{tancik2022block, tao2024silvr}, and artifacts from sparse observations\cite{wu2025sparse2dgs}. However, when view sparsity is high, artifacts can still occur even with techniques like geometric regularization \cite{younis2025sparse}. Therefore, we propose a method to address artifacts on the sparsely-observed floor by representing it as a low-frequency texture, ensuring that the agent perceives the floor clearly, without artifacts.

\begin{figure*}[!ht]
    \centering
    \includegraphics[width=0.95\linewidth]{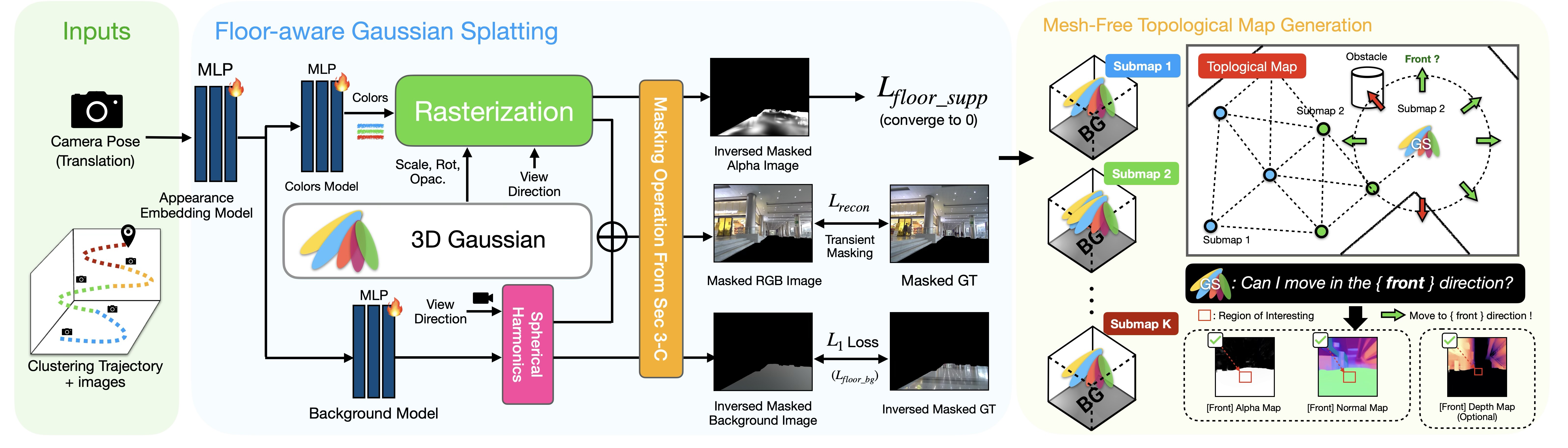}
    \caption{An overview of the NVSim framework. Given an RGB image sequence and camera poses, we first cluster the trajectory into submaps. For each submap, we generate robust floor masks using our hybrid segmentation method and then reconstruct the scene with Floor-Aware Gaussian Splatting. Finally, a mesh-free topological map is automatically generated from this collection of geometric cues such as alpha map and surface normals.}
    \label{fig:fig_framework}
\end{figure*}

\section{Novel View Synthesis Simulator}
\subsection{Problem Formulation}

Our goal is to automatically construct a simulator for a large-scale indoor environment, enabling agent navigation, using only an image sequence $\mathcal{I}$ and its corresponding camera poses $\mathcal{P}$ collected during a robot traversal. Given a set of RGB images $\mathcal{I} = \{I_i\}_{i=1}^{N}$, camera poses $\mathcal{P} = \{P_i\}_{i=1}^{N}$ where $P_i \in \mathbb{R}^{4 \times 4}$, and camera intrinsics $K$, our system generates a navigable topological map $\mathcal{G} = (\mathcal{V}, \mathcal{E})$. Here, $\mathcal{V}$ represents a set of reachable viewpoint nodes for the agent, and $\mathcal{E}$ denotes the set of collision-free edges connecting them. The core idea is to first learn a 3D scene representation of the environment and then automatically construct the graph $\mathcal{G}$ by identifying traversable paths within this representation.

To achieve this, we decompose the problem into two main stages: (1) learning a 3D representation for large-scale scenes and (2) constructing a topological map for navigation.

The first stage involves learning a mapping function $\Phi_\theta$ that generates a scene representation $S$ from the given images and poses:
\begin{equation}
S = \Phi_{\theta}(P, K), \text{where } \theta \text{ is learned from } (\mathcal{I}, \mathcal{P}, K).
\end{equation}
However, this process presents two significant challenges:

\textbf{Challenge 1 (Scalability):} Representing an entire large-scale indoor environment with a single model $\Phi_\theta$ leads to capacity bottlenecks and degraded reconstruction quality.

\textbf{Challenge 2 (Reconstruction Fidelity for Navigation):} Robot traversal data is often forward-facing. This leads to visual artifacts during reconstruction, especially in sparsely observed areas such as floors. These artifacts compromise the reliability of navigable space and make the representation unsuitable for a navigation simulator.

The second stage is to construct the graph $\mathcal{G}$, composed of viewpoint nodes $\mathcal{V}$ and edges $\mathcal{E}$, from the scene representation $S$. This stage introduces its own key challenge:

\textbf{Challenge 3 (Mesh-free Traversability Checking):} Given a mesh-free scene representation $S$, we aim to automatically discover navigable space and construct a graph by checking traversability from the rendered scene $S$, without relying on traditional collision checking mechanisms. In the following sections, we will address each of these challenges in detail.

\subsection{Large-Scale Scene Decomposition}
\label{sec:submapping}
Training a single 3D scene representation model $\Phi_\theta$ for an entire large-scale indoor environment is challenging due to capacity bottlenecks and high computational costs.
To address this, we adopt a divide-and-conquer strategy that partitions the scene into a set of submaps, inspired by prior works \cite{tancik2022block, tao2024silvr}. Specifically, we apply Agglomerative Clustering to the camera trajectory $\{\boldsymbol{t}_i\}_{i=1}^N \subset \mathbb{R}^3$, to segment the environment into $C$ submaps.

However, a naive partitioning often leads to degraded visual quality at the boundaries between submaps where view diversity is limited. To mitigate this, we introduce an overlapping strategy. Each submap is augmented with images from its spatially adjacent neighbors that fall within a distance threshold $\delta$. This data overlap ensures that submap boundaries are reconstructed with rich multi-view information, promoting seamless transitions and overall geometric consistency. This approach allows us to overcome memory constraints and enables each smaller model, $\Phi_{\theta_c}$, to capture more fine-grained details within its designated area. These individually trained models are then merged for rendering as described in Section~\ref{subsec:navigable_topomap}.

\subsection{Floor-Aware Gaussian Splatting}
As noted in Challenge 2, image data from robot traversals is inherently forward-biased due to the fixed camera perspective.  This leads to significant visual artifacts in 3D Gaussian representation $\Phi_{\theta}$, such as floaters and blurry patches on the floor plane, which severely compromise the navigable space.

We introduce a floor-aware reconstruction method that prevents the formation of unstable 3D Gaussians in the floor region. Our core idea is to replace the unstable Gaussian modeling of the floor. We leverage a background MLP based on Spherical Harmonics for rendering a clean ground plane. We detail this approach in Section~\ref{sec:hybrid_floor} and Section~\ref{sec:mask_guided}.

\subsubsection{\textbf{Hybrid Floor Segmentation}}
\label{sec:hybrid_floor}
\begin{figure}[t]
    \centering
    \includegraphics[width=0.95\linewidth]{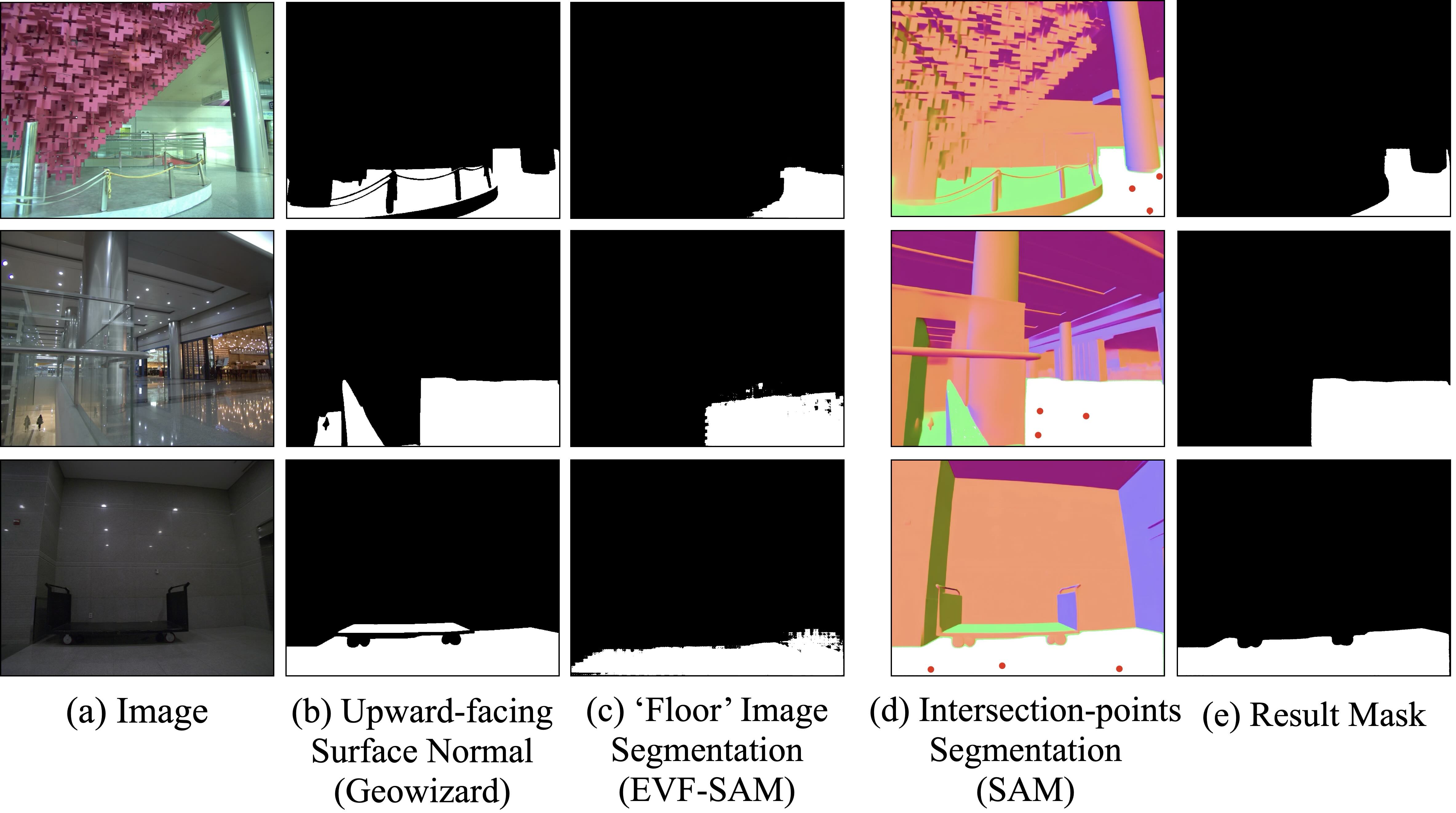}
    \caption{Results comparing the masks from the hybrid floor segmentation process; (d) shows red points sampled from $M_{\text{cand}}$.}
    \label{fig:floor_masking}
\end{figure}
The first step in training $\Phi_{\theta}$ is to identify the floor region in the training images. To achieve this, we leverage vision foundation models for their strong scene understanding capabilities. However, a single source of information is often insufficient. As shown in Fig.~\ref{fig:floor_masking} (b) and (c), Semantic segmentation models can be prone to challenging lighting reflections or textures, While surface normal estimation classifies all flat surfaces as the floor, it cannot yield floor information exclusively. Therefore, we propose a hybrid segmentation method that fuses both semantic and geometric cues to generate a robust floor mask.

First, we generate two types of candidate masks in parallel from each training RGB image $I_i \in \mathcal{I}$. The first mask, $M_{sem} \in \{0, 1\}$, is produced by EVF-SAM, a zero-shot semantic segmentation model, using the text prompt "floor". While this mask is semantically accurate, we observed that it sometimes fails to capture the entire floor area.

The second mask, $M_{norm} \in \{0, 1\}$, is derived from geometric information. We use the GeoWizard model, a zero-shot surface normal estimator, to compute the surface normal map $N$. We then create the mask by selecting pixels corresponding to horizontal surfaces. This mask reliably includes all geometrically flat areas, but its drawback is the inclusion of non-floor surfaces like desks and shelves.

Next, we extract a set of high-confidence floor candidates, $M_{cand}$, by computing the intersection of these two masks:
\begin{equation}
    M_{cand} = M_{sem} \cap M_{norm}.
\end{equation}
This intersection contains only the pixels that are both semantically classified as "floor" and geometrically identified as "flat". To handle potential noise, we apply a connected component analysis to $M_{cand}$ and remove any isolated pixel groups smaller than a predefined area threshold. This step ensures that only large, contiguous floor regions remain as final candidates.

Finally, we generate the final floor mask $M_{final}$ from these refined candidates. We sample a small set of point coordinates $\mathcal{X} = \{ \boldsymbol{x}_k \}_{k=1}^K$ from the $M_{cand}$ region, where we empirically set $K=3$. As shown in Fig.~\ref{fig:floor_masking} (d), we then use these points as prompts for the SAM2 model to infer the final mask. Crucially, we use the surface normal map $N$ as the input to SAM2 instead of the original RGB image:
\begin{equation}
    M_{final} = \text{SAM2}(N, \mathcal{X}).
\end{equation}
This approach provides two key advantages. First, the surface normal map encodes the intrinsic 3D structure of the scene and is invariant to photometric variations like lighting changes or complex textures present in the RGB image. This allows SAM2 to more faithfully track the physical boundaries of objects, resulting in more accurate segmentation. Second, even with only a few point prompts from a high-confidence region, SAM2 can consistently segment the entire floor area in the surface normal map, effectively recovering regions that may have been missed by $M_{sem}$.

\subsubsection{\textbf{Floor-Aware Reconstruction}}
\label{sec:mask_guided}
Our core idea is to prevent the creation of unstable 3D Gaussians in the floor region, which suffers from insufficient view diversity. Instead, we task a spherical harmonics-based background MLP model with representing the floor. This approach is inspired by Splatfacto-W\cite{xu2024splatfacto}, which uses a three-layer background MLP with appearance embeddings to model the sky in outdoor scenes. We adapt this concept and specialize it to solve the specific problem of representing static floor planes in indoor environments. To this end, we design two loss functions for the floor region.

\textbf{Floor Suppression Loss}($L_{floor\_supp}$) explicitly penalizes the formation of 3D Gaussians in areas masked as the floor ($p \in M_{final}$). It applies a penalty to drive the accumulated alpha value $\alpha$ of any rasterized Gaussians in this region towards zero:
\begin{equation}
    L_{floor\_supp} = \left\| M_{final} \odot \alpha \right\|_{1}
\end{equation}

\textbf{Background Floor Loss} ($L_{floor\_bg}$) guides the background model to consistently represent the floor across different views. It minimizes the difference between the ground-truth image $\mathcal{I}$ and the background model's prediction $\hat{I}_{BG}$ within the masked floor area. This encourages the model to learn the overall color and low-frequency texture of the floor:
\begin{equation}
    L_{floor\_bg} = \left\| M_{final} \odot \left( \hat{I}_{BG} - \mathcal{I} \right) \right\|_{1}
\end{equation}

Our total loss function combines a standard reconstruction loss for the non-masked areas ($L_{L1} + L_{SSIM}$) with our two proposed floor losses. We also incorporate the robust masking technique from Splatfacto-W to handle transient objects.
\begin{equation}
    L_{recon} = (1 - \lambda_{SSIM}) \cdot L_{1} \;+\; \lambda_{SSIM} \cdot L_{SSIM}
\end{equation}
\begin{equation}
    L_{total} = L_{recon} + \lambda_{supp} \cdot L_{floor\_supp} + \lambda_{bg} \cdot L_{floor\_bg}
\end{equation}
We set $\lambda_{SSIM}=0.2$, and both $\lambda_{supp}$ and $\lambda_{bg}$ to 1.0. We note that while our baseline, Splatfacto-W, introduces an Alpha Loss to prune Gaussians in regions represented by the background model, our experiments show that omitting this loss yields better results when using our proposed losses.

The final rendering process for $\Phi_\theta$ follows the standard alpha blending of the 2D background model and the Gaussian rasterization output. The rasterization process computes a color value $\hat{I}_{GS}$ and an accumulated alpha $\alpha$ for each pixel. The background model, based on spherical harmonics, represents texture-less surfaces, distant scenery, and, in our case, the floor. The final rendered image $\hat{I}$ is computed as:
\begin{equation}
    \hat{I} = \hat{I}_{GS} + (1 - \alpha) \odot \hat{I}_{BG}
\end{equation}
This method allows us to perform 3D scene representation that reliably removes floor artifacts from image sequences typically collected by robot traversals.

\subsubsection{\textbf{Camera Pose Embedding}}
The baseline Splatfacto-W conditions its appearance embedding on a discrete index, one for each training camera. While effective for seen views, this approach poses a challenge for novel viewpoints, since no such index exists for them and they must fall back on heuristics for appearance generation.

To overcome this, we employ a learnable MLP that directly maps the continuous camera's 3D position ($\mathbf{t} \in \mathbb{R}^3$) to an appearance embedding. This enables spatially continuous inference, allowing for plausible interpolation of appearances for novel views.

While including camera rotation as an input could provide direction-aware information, we found it led to color inconsistencies when rendering multiple views for the topological map. Therefore, we use only the camera position to ensure consistent appearance rendering during map construction.

\subsection{Mesh-free Topological Map Generation}
\label{subsec:navigable_topomap}

\begin{algorithm}[t]
\caption{Mesh-free Topological Map Generation}
\label{alg:topomap_generation}
\SetKwInput{KwInput}{Input}
\SetKwInput{KwOutput}{Output}
\DontPrintSemicolon

\KwInput{Initial viewpoint $v_0$, Set of submap models $\{\Phi_\theta\}$}
\KwOutput{Navigable Topological Map $\mathcal{G} = (\mathcal{V}, \mathcal{E})$}

Initialize $\mathcal{G}=(\{v_0\}, \emptyset)$, $Q=\{v_0\}$\;
\While{$Q$ is not empty}{
    $v_{curr} \leftarrow Q.\text{pop}()$\;
    Render spherical views $\{\Omega_{rgb}, \Omega_{normal}, \Omega_{alpha}\}$ from the closest submap model $\Phi_{\theta_c}$ for $v_{curr}$\;
    \ForEach{direction $d$ in 8 neighbors}{
        \If{$\mathtt{ROI}(\Omega_{alpha_d}) < \tau_{alpha}$ \text{ and } $\mathtt{ROI}(\Omega_{normal_d}) > \tau_{normal}$}{
            $v_{next} \leftarrow \mathtt{NextViewpoint}(v_{curr}, d, \tau_{dist})$\;
            Add $v_{next}$ to $\mathcal{V}$ (if new) and $Q$\;
            Add edge $(v_{curr}, v_{next})$ to $\mathcal{E}$\;
        }
    }
}
\KwRet{$\mathcal{G}$}\;
\end{algorithm}
As outlined in Challenge 3, a core objective of our work is to automatically generate a navigable topological map $\mathcal{G} = (\mathcal{V}, \mathcal{E})$ from the learned static 3D Gaussian scene representation $S$. Unlike conventional simulators that rely on explicit meshes for collision detection \cite{savva2019habitat,shen2021igibson}, 3DGS does not provide such a structure. Therefore, we propose an algorithm that directly explores traversable free space from the ensemble of learned submap models $\{\Phi_{\theta_c}\}$ and merges them into a unified navigable graph.

Our algorithm is based on Breadth-First Search (BFS) and incrementally expands the navigable space starting from an arbitrary viewpoint $v_0 \in \mathcal{V}$ on the training trajectory. As detailed in Algorithm~\ref{alg:topomap_generation}, for a given viewpoint $v_{curr}$, we first select the scene model $\Phi_{\theta_c}$ corresponding to the closest submap cluster based on spatial distance \textbf{(Line 4)}. We then use the selected model $\Phi_{\theta_c}$ to render a spherical image and its accumulated alpha map, and employ a vision foundation model to infer the corresponding surface normal map. The rendered spherical image serves as the local scene representation $S$ at the current viewpoint.

From $v_{curr}$, we check for traversability in eight cardinal directions. We determine a path to be traversable if the average values of the accumulated alpha ($\Omega_{alpha_d}$) and the upward-facing component of the surface normal ($\Omega_{normal_d}$) within a predefined Region of Interest (RoI) both satisfy their respective thresholds \textbf{(Line 6)}. If both conditions are met, we place a new viewpoint $v_{next}$ at a fixed distance $\tau_{dist}$ along that direction \textbf{(Line 7)}. A smaller $\tau_{dist}$ results in smoother, more continuous navigation, while a larger value leads to more discrete movements. If $v_{next}$ already exists, we simply establish an edge to it from $v_{curr}$; otherwise, we add it to the map as a new viewpoint. \textbf{(Line 8)}

We set the height $z$ of the new viewpoint to match that of the nearest point on the original trajectory $\mathcal{P}$. This approach incorporates real-world height information, which the flattened floor representation $S$ cannot provide, thereby ensuring stable agent movement. The search terminates when no more new viewpoints can be added. We empirically set $\tau_{alpha}=0.95$, $\tau_{normal}=0.85$, and $\tau_{dist}=2.5m$.

Since all submaps share extrinsic parameters defined in a common coordinate system, the absolute coordinates $\mu_\theta$ of all Gaussians remain globally consistent, even though each submap model is trained independently. This ensures both visual and geometric continuity when transitioning between submaps during the map generation process.

We also considered alternative methods for obstacle detection. Using rendered depth maps from 3DGS was found to be unreliable, as the 2D background model often represents texture-less planes as empty space. While zero-shot metric depth estimation on spherical images is an option, we did not adopt it due to the known issue of depth discontinuities at the equirectangular boundaries \cite{saura2021spherical}.

\section{Experimental Setup}
\textbf{Scene Representation Baselines.} We select 3D Gaussian Splatting (3DGS) \cite{kerbl2023gaussiansplatting} as our core baseline, given its notable advancements in scene representation. 3DGS not only offers high rendering quality but also provides the real-time rendering speeds and fast training times. These characteristics make it a more suitable foundation for our research compared to earlier NeRF-based approaches.
To evaluate the scene representation quality, we compare our proposed method, Splatfacto-i, against two baselines from the official Nerfstudio framework. The first is Splatfacto, the standard 3DGS implementation, and the second is Splatfacto-W \cite{xu2024splatfacto}, which is designed to handle dynamic objects.

\textbf{Datasets.} We conduct our experiment on the COEX dataset from the Large-Scale Indoor Localization datasets \cite{lee2021large}. This large-scale public indoor environment spans approximately 6,000 m² and includes diverse areas such as corridors, lobbies, and commercial areas. The dataset features challenging real-world factors for 3D reconstruction, including dynamic crowds, reflective surfaces, varied textures, and numerous signages. These characteristics make it an ideal testbed for evaluating our scene representation and its application in navigation tasks within the our framework. The dataset also provides ground-truth camera poses, ensuring high geometric fidelity for our reconstruction. We initialize our Gaussian Splatting models using the point clouds from the provided LiDAR data. To handle the large-scale environment, we decompose the COEX scene into 15 trajectory-based submaps, as described in Section~\ref{sec:submapping}. We then train a separate scene model $\Phi_{\theta_c}$ for each submap. All experimental results reported below represent the average performance across these 15 scene models.

\section{Results}
In this section, we aim to address the following questions:
\begin{enumerate}[leftmargin=*]
\item Does Floor-Aware Gaussian Splatting effectively remove artifacts to provide scene representations for navigation?
\item Can Mesh-free Topological Map Generation construct valid large-scale paths?
\item Do existing zero-shot and trained R2R methods work in the large-scale NVSim environment?
\end{enumerate}
\subsection{3D Scene Representation}
\begin{figure}[t]
    \centering
    \includegraphics[width=0.95\linewidth]{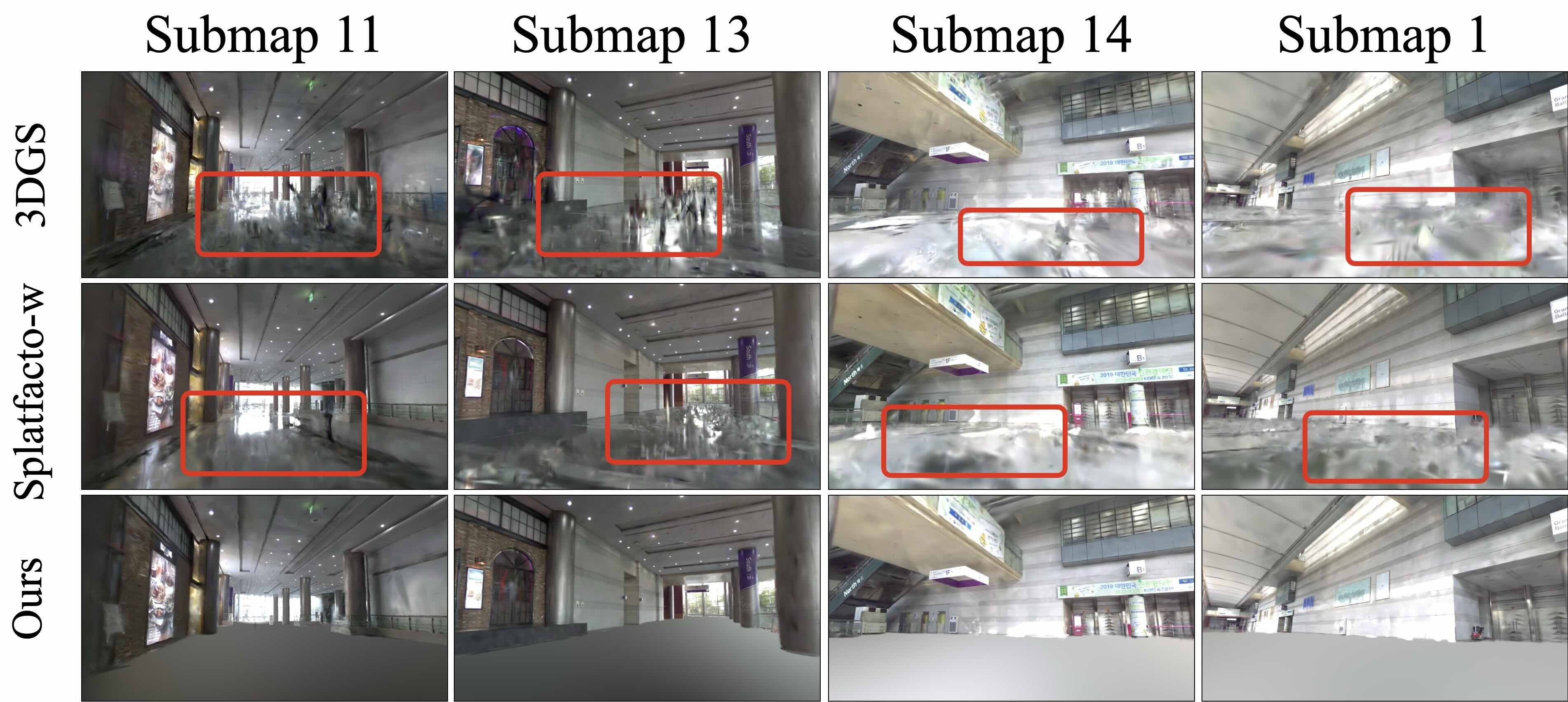}
    \caption{Quantitative comparison of novel view scene representations across different submaps.}
    \label{fig:recon_comparative_novelview}
\end{figure}

\begin{figure}[t]
    \centering
    \includegraphics[width=0.95\linewidth]{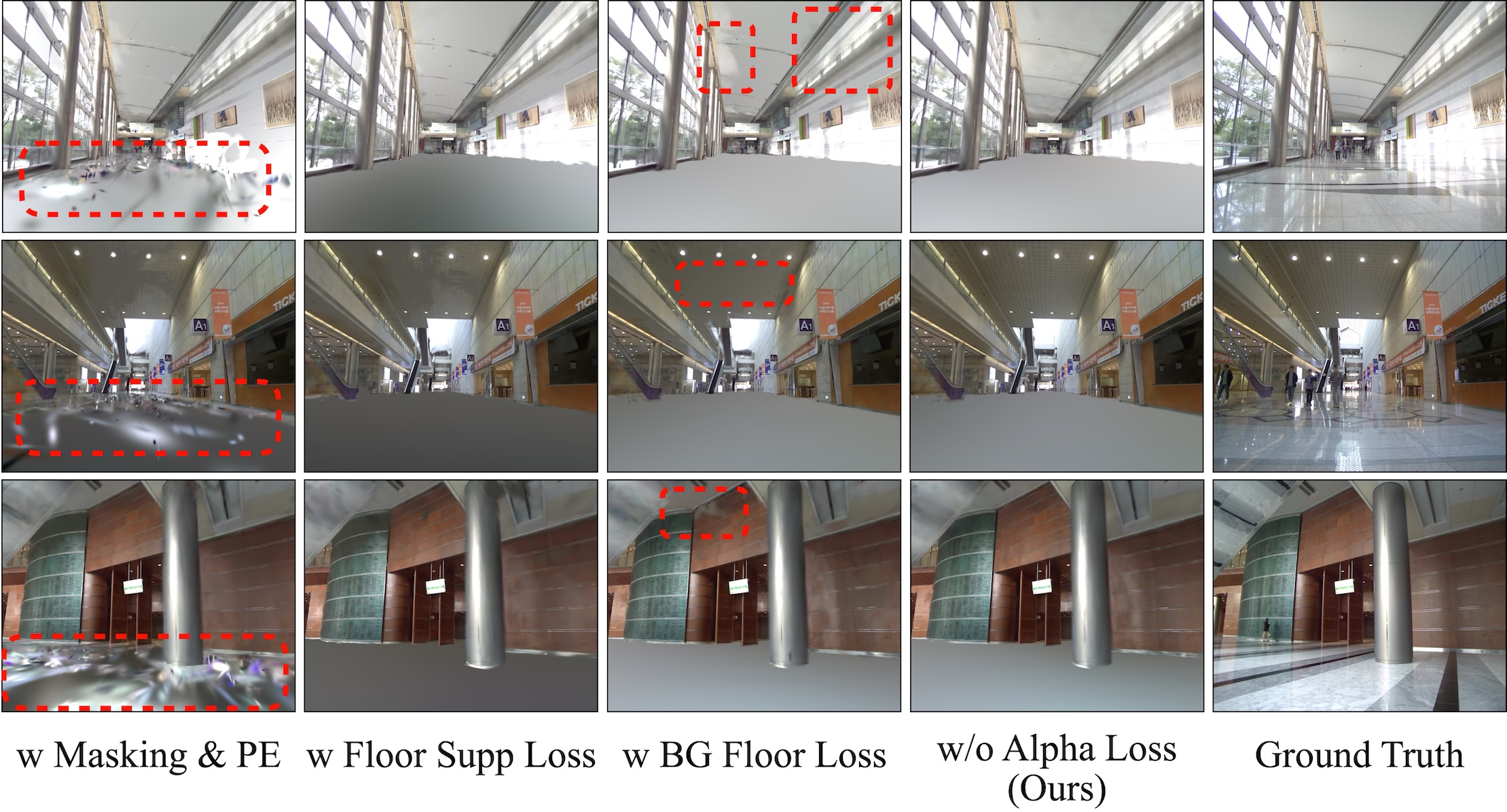}
    \caption{Qualitative results from the ablation study on our navigation-specific scene representation.}
    \label{fig:recon_ablation}
\end{figure}

We evaluate our method against a baseline 3D scene representation across 15 submaps, reporting the average Peak Signal-to-Noise Ratio (PSNR), Structural Similarity Index (SSIM), and Learned Perceptual Image Patch Similarity (LPIPS). To specifically assess the representational capacity of our Gaussians and the Background Model, we conduct a comparative analysis on scenes with and without the floor region. As shown in Table~\ref{tab:performance_combined}, our method achieves a marginal improvement over the baseline while successfully incorporating navigation-specific features. Furthermore, as illustrated in Fig.~\ref{fig:recon_comparative_novelview}, our approach effectively eliminates floor artifacts in novel views, rendering a distinct boundary between walls and the ground, which is a notable improvement over the baseline. Unlike methods that average embedding vectors to infer appearance, our approach leverages the camera pose to infer appearance embeddings. This allows for a more accurate representation of lighting reflections on the floor, even in novel views.

\begin{table}[t]
\centering
\begin{threeparttable}
\caption{Quantitative and Qualitative Comparison of Scene Representation Models}
\label{tab:performance_combined}
\setlength{\tabcolsep}{5pt}
\begin{tabular}{l|ccc|ccc}
\toprule
\textbf{Model} & \textbf{PSNR ↑} & \textbf{SSIM ↑} & \textbf{LPIPS ↓} & \textbf{VD} & \textbf{TO} & \textbf{FA} \\
\midrule
\rowcolor{gray!20} \multicolumn{7}{l}{\textbf{With Floor}} \\
3DGS   & 20.948 & 0.816 & 0.403 & \checkmark &            &            \\
Splatfacto-w & \textbf{21.460} & 0.823 & \textbf{0.385} &            & \checkmark &            \\
\textbf{Splatfacto-i (Ours)} & 20.217 & \textbf{0.824} & 0.399 & \checkmark & \checkmark & \checkmark \\
\midrule
\rowcolor{gray!20} \multicolumn{7}{l}{\textbf{Without Floor*}} \\
3DGS   & 23.362 & 0.874 & 0.257 & \checkmark &            &            \\
Splatfacto-w & 23.718 & 0.877 & 0.242 &            & \checkmark &            \\
\textbf{Splatfacto-i (Ours)} & \textbf{23.760} & \textbf{0.883} & \textbf{0.234} & \checkmark & \checkmark & \checkmark \\
\bottomrule
\end{tabular}
\begin{tablenotes}[flushleft]\footnotesize
\item VD = View-dependent Effects,\quad TO = Transient Objects Handling,\quad FA = Floor Artifacts Handling.
\item * : Evaluated on non-floor regions.
\end{tablenotes}
\end{threeparttable}
\end{table}

Additionally, our 2D Background Model renders the floor with location-aware colors that are faithful to the original surface, seamlessly integrating with the Gaussian representation to produce a natural-looking floor area.
Table~\ref{tab:comparative} and Fig.~\ref{fig:recon_ablation} present an ablation study of our proposed Floor-Aware Gaussian Splatting. We observe that simply masking the floor region in training images degrades rendering performance and fails to effectively remove floor artifacts. In contrast, introducing our proposed Floor Suppression Loss successfully eliminates these artifacts. The addition of the Background Floor Loss further enables the representation of lighting-dependent color variations on the floor, which is corroborated by the performance increase in the "With Floor" metrics in Table~\ref{tab:comparative}. Finally, we find that removing the Alpha Loss from Splatfacto-w improves both the quantitative metrics and the visual quality. As seen in the figure, this allows for a more detailed representation of texture-less surfaces with Gaussians. This suggests that the Alpha Loss, by overly suppressing Gaussian representation, is not well-suited for indoor environments that contain large surfaces like ceilings and walls.

\begin{table}[t]
\centering
\begin{threeparttable}
\caption{Ablation Study of Splatfacto-i, where each component is added cumulatively to the baseline}
\label{tab:comparative}
\footnotesize 
\setlength{\tabcolsep}{2pt}
\begin{tabular}{l|ccc|ccc} 
\toprule
\multirow{2}{*}{\textbf{Experiment}} & 
\multicolumn{3}{c|}{\textbf{With Floor}} &
\multicolumn{3}{c}{\textbf{Without Floor}} \\
& PSNR$\uparrow$ & SSIM$\uparrow$ & LPIPS$\downarrow$
    & PSNR$\uparrow$ & SSIM$\uparrow$ & LPIPS$\downarrow$ \\
\midrule
Splatfacto-w (Base) & \textbf{21.46} & 0.823 & \textbf{0.385} & 23.71 & 0.877 & 0.242  \\
w Masking \& PE & 18.37 & 0.800 & 0.430   & 23.60   & 0.879 & 0.246   \\
w Floor Supp Loss & 18.17 & 0.806 & 0.429   & 23.36 & 0.877 & 0.250  \\
w BG Floor Loss  & 20.07  & 0.819 & 0.412   & 23.41  & 0.878 & 0.247   \\
\rowcolor{gray!15}
\textbf{w/o Alpha Loss (Ours)}  & 20.21 & \textbf{0.824} & 0.399 & \textbf{23.76} & \textbf{0.883} & \textbf{0.234}   \\
\bottomrule
\end{tabular}
\begin{tablenotes}[flushleft]\footnotesize
    \item PE : Camera Pose Embedding Instead of a Camera Index.
    \end{tablenotes}
\end{threeparttable}
\end{table}

\begin{table}[t]
\centering
\caption{Results of Pairwise Comparison using Multimodal Large Language Models (mLLMs). Total 375 Pairs are used for each comparison.}
\label{tab:pairwise_comparison}
\setlength{\tabcolsep}{6pt}
\renewcommand{\arraystretch}{1.1}
\scriptsize 
\begin{tabular}{l|c|c}
\toprule
\textbf{Pair Comparison (Winner Bold)} & \textbf{Win Points (\%)} & \textbf{Tie Rate (\%)} \\
\midrule
\rowcolor{gray!20} \multicolumn{3}{l}{\textbf{Gemini-2.5-Flash}} \\
\textbf{Splatfacto-i (Ours)} vs. Splatfacto-w & 77\% \textcolor{gray}{(290)} & 21\% \textcolor{gray}{(80)} \\
\textbf{Splatfacto-i (Ours)} vs. 3DGS        & 88\% \textcolor{gray}{(333)} & 14\% \textcolor{gray}{(54)} \\
\textbf{Splatfacto-w} vs. 3DGS               & 78\% \textcolor{gray}{(296)} & 23\% \textcolor{gray}{(88)} \\
\midrule
\rowcolor{gray!20} \multicolumn{3}{l}{\textbf{GPT-5-mini-2025-08-07}} \\
\textbf{Splatfacto-i (Ours)} vs. Splatfacto-w & 76\% \textcolor{gray}{(286)} & 27\% \textcolor{gray}{(104)} \\
\textbf{Splatfacto-i (Ours)} vs. 3DGS        & 85\% \textcolor{gray}{(320)} & 19\% \textcolor{gray}{(72)} \\
\textbf{Splatfacto-w} vs. 3DGS               & 65\% \textcolor{gray}{(246.5)} & 65\% \textcolor{gray}{(245)} \\
\midrule
\rowcolor{gray!20} \multicolumn{3}{l}{\textbf{GPT-5-2025-08-07}} \\
\textbf{Splatfacto-i (Ours)} vs. Splatfacto-w & 93\% \textcolor{gray}{(350)} & 8\% \textcolor{gray}{(32)} \\
\textbf{Splatfacto-i (Ours)} vs. 3DGS        & 95\% \textcolor{gray}{(356.5)} & 7\% \textcolor{gray}{(27)} \\
\textbf{Splatfacto-w} vs. 3DGS               & 83\% \textcolor{gray}{(313)} & 15\% \textcolor{gray}{(56)} \\
\bottomrule
\end{tabular}
\end{table}

\begin{figure}[t]
    \centering
    \includegraphics[width=0.9\linewidth]{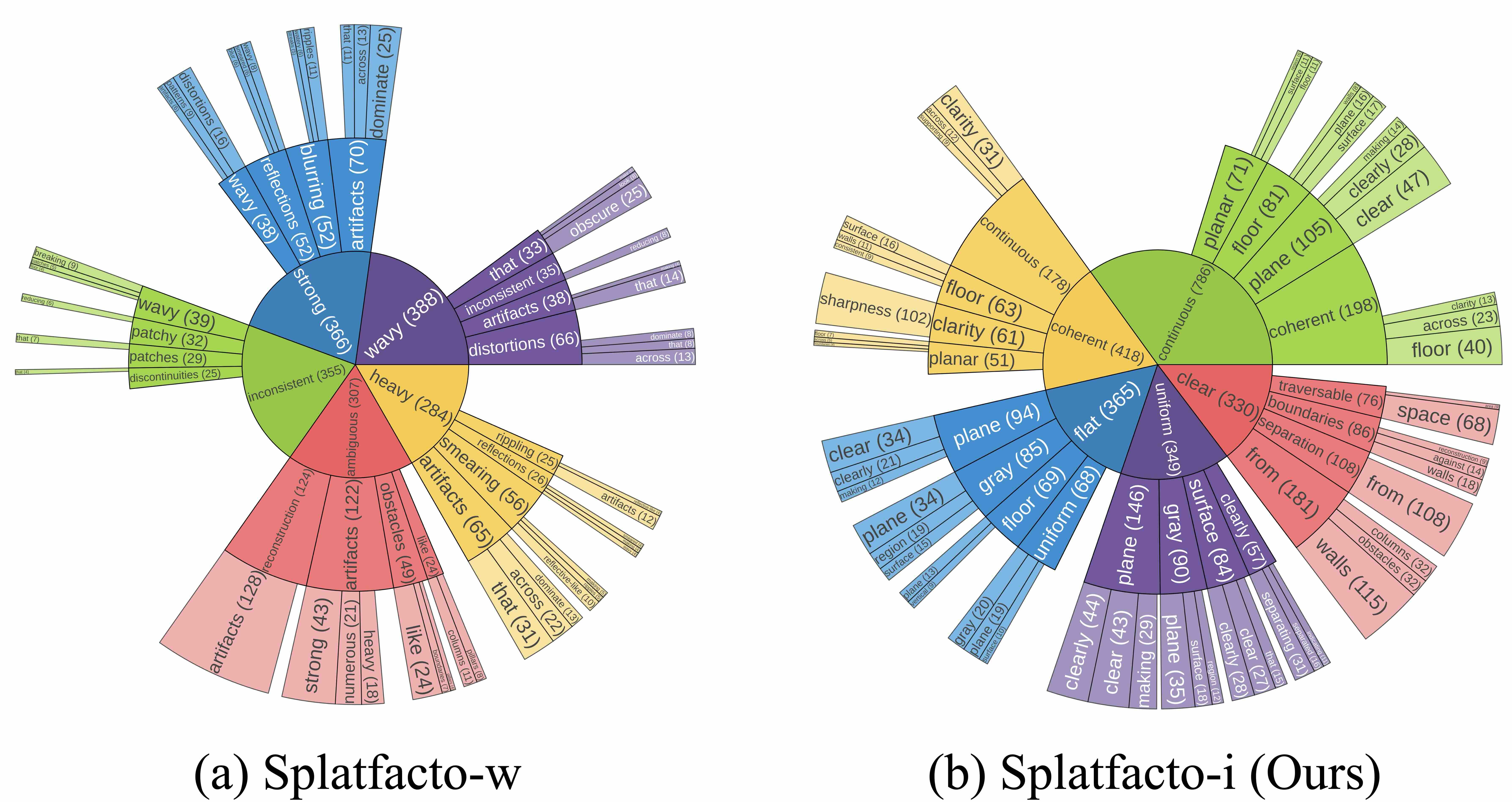}
    \caption{Visualization of mLLM reasoning for pairwise comparison between our method and baseline scene representation images}
    \label{fig:recon_mllm_iqa}
\end{figure}

The primary objective of our scene representation is to provide high-quality visual inputs for Vision-Language Navigation (VLN) tasks. As recent VLN research shifts towards zero-shot methods leveraging Multi-modal Large Language Models (mLLMs) \cite{zhou2024navgpt,chen2024mapgpt, qiao2025open}, it has become crucial to supply these foundation models with scenes that clearly separate navigable floors from walls for effective spatial reasoning. To this end, we conduct a pairwise comparison using mLLMs to demonstrate the advantage of our Floor-Aware Gaussian Splatting over baseline methods. In this setup, an mLLM evaluates two images rendered from the same viewpoint and selects the one it prefers \cite{gu2024survey}.
We compare three representations (Ours, Splatfacto-w, and 3DGS) using a set of 5 seen-view and 20 novel-view images from each of our 15 submaps. We task three distinct mLLMs with this preference selection in the navigation context. To ensure fairness, we present each image pair twice with the order swapped. A representation scores 1 point for a consistent win and 0.5 points for a tie.
The results, summarized in Table~\ref{tab:pairwise_comparison}, show that our method is consistently preferred over both baselines. While Splatfacto-w scores slightly higher than 3DGS, the high tie rate between them indicates no strong preference. Fig.~\ref{fig:recon_mllm_iqa} visualizes the mLLMs' qualitative reasoning as a Sunburst Chart. The analysis reveals a consistent pattern: mLLMs criticize baseline methods for "floor artifacts" that hinder navigation, whereas they praise our method for its "consistent floor surface." This experiment shows that our method effectively removes artifacts that foundation models perceive as detrimental to spatial understanding.

\subsection{Mesh-Free Topological Map Generation}
\begin{figure}[t]
    \centering
    \includegraphics[width=0.7\linewidth]{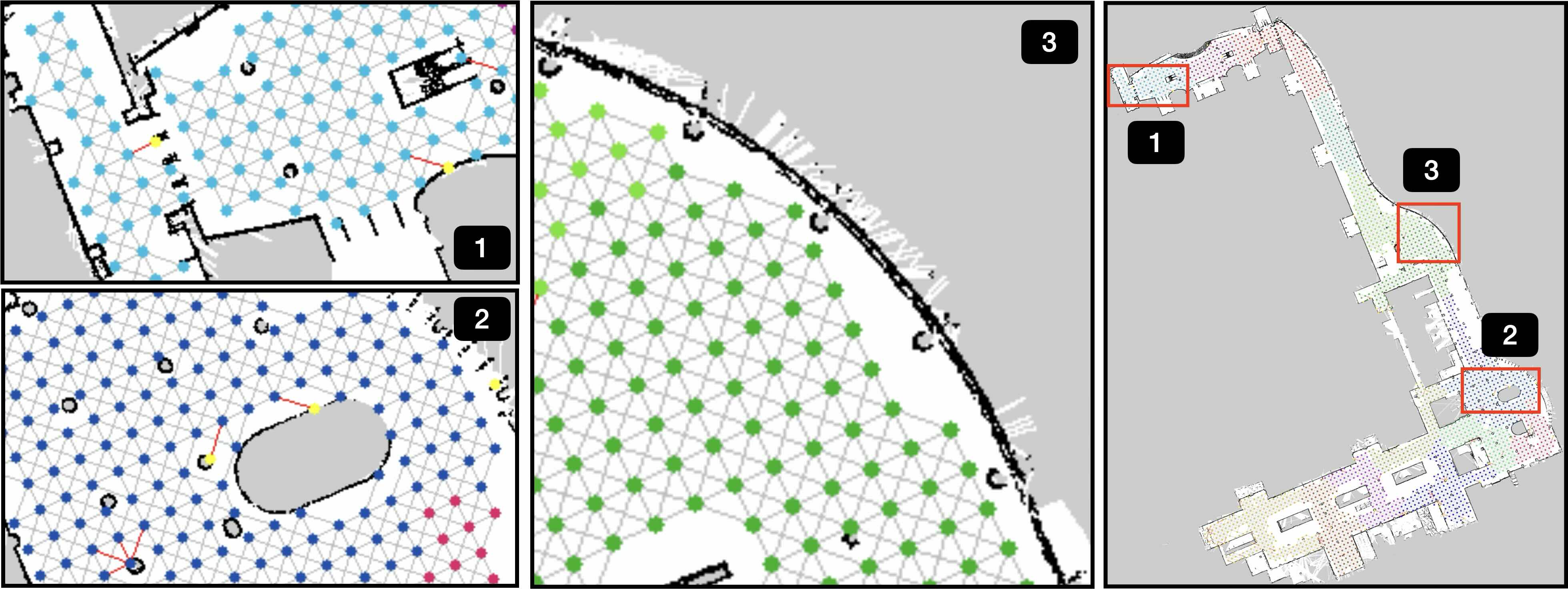}
    \caption{Quantitative evaluation of our topological map on the 2D map by OctoMap. Novel viewpoints are shown with nodes and edges; node colors denote submap indices, and yellow nodes and red edges mark invalid cases.}
    \label{fig:fig_ros}
\end{figure}

This section evaluates the validity of topological maps generated from different scene representations. We quantitatively assess our method's ability to generate a valid topological map against two baselines: 3DGS and Splatfacto-w. We define a common Region of Interest (ROI) across all baseline models. For ground truth, we generate a 2D occupancy map using the OctoMap framework \cite{hornung2013octomap} (Fig.~\ref{fig:fig_ros}), using LiDAR scans and camera extrinsics from the dataset. 

As shown in Table \ref{tab:valid_node_edge_ratios}, we assess the validity of the generated topological map's nodes and edges based on two criteria:
\textbf{Node Validity}, a sampled node is valid if it falls within a traversable region of the 2D occupancy map.
\textbf{Edge Validity}, an edge is valid if it connects two valid nodes and the straight path between them does not cross any obstacles.

\begin{figure}[t]
    \centering
    \includegraphics[width=0.9\linewidth]{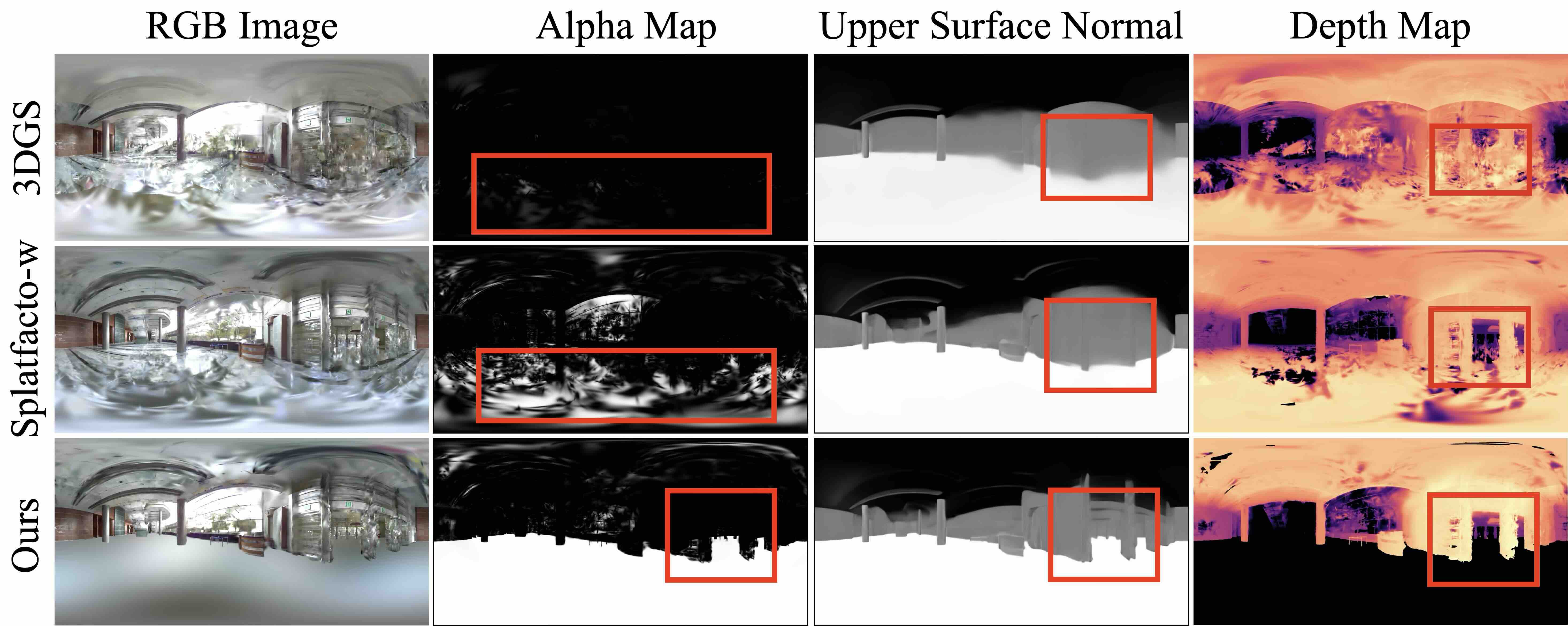}
    \caption{Qualitative comparison of rendered scenes, depth maps, and estimated upper-surface normals across different scene representation models.}
    \label{fig:recon_quantitative}
\end{figure}

We assumed that in the alpha map, the absence of Gaussians on the floor plane would indicate traversable space. However, as shown in Fig.~\ref{fig:recon_quantitative}, this assumption is not valid for baselines like 3DGS and Splatfacto-w, which explicitly reconstruct the floor surface with Gaussians. Consequently, we did not evaluate these methods using our Alpha map approach.

\begin{figure}[t]
    \centering
    \includegraphics[width=0.8\linewidth]{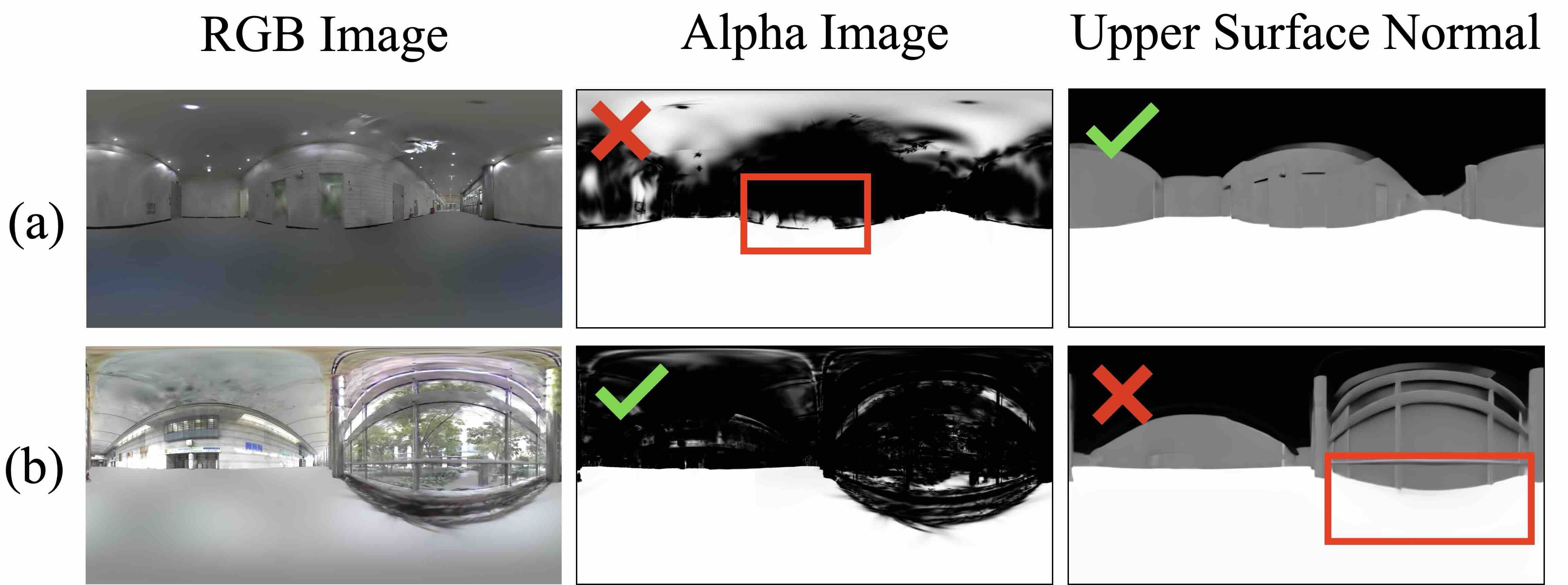}
    \caption{Illustration of failure cases in mesh-free traversability checking using Alpha Maps and upper-surface normals}

    \label{fig:why_ensamble_is_important}
\end{figure}

\begin{table}[t]
\centering
\begin{threeparttable}
\caption{Valid node and edge ratios (\%) of the topological map evaluated on the 2D map.}
\label{tab:valid_node_edge_ratios}
\setlength{\tabcolsep}{3pt} 
\scriptsize
\renewcommand{\arraystretch}{0} 
\begin{tabular}{l|l|cccc}
\toprule
& \textbf{Model} & \textbf{Alpha\&Normal} & \textbf{Alpha} & \textbf{Normal} & \textbf{Depth} \\
\midrule
\multirowcell{3}[-1.5ex][c]{\rotatebox[origin=c]{90}{\textbf{Node}}} &
3DGS   & - & - & \makecell{90.3*\\ \textcolor{gray!80}{(252/279)}} & \makecell{93.6* \\ \textcolor{gray!80}{(220/235)}} \\
& Splatfacto-w & - & - & \makecell{88.8* \\ \textcolor{gray!80}{(286/322)}} & \makecell{91.9 \\ \textcolor{gray!80}{(2208/2402)}} \\
& \textbf{Splatfacto-i (Ours)}  & \makecell{\textbf{97.9} \\ \textcolor{gray!80}{(\textbf{2050}/2095)}} & \makecell{\textbf{96.0} \\ \textcolor{gray!80}{(\textbf{2053}/2139)}} & \makecell{\textbf{92.4} \\ \textcolor{gray!80}{(\textbf{2253}/2438)}} & \makecell{\textbf{92.9} \\ \textcolor{gray!80}{(\textbf{2184}/2352)}} \\
\midrule
\multirowcell{3}[-1.5ex][c]{\rotatebox[origin=c]{90}{\textbf{Edge}}} & 
3DGS   & - & - & \makecell{84.9* \\ \textcolor{gray!80}{(741/873)}} & \makecell{91.7* \\ \textcolor{gray!80}{(676/737)}} \\
& Splatfacto-w & - & - & \makecell{83.3* \\ \textcolor{gray!80}{(854/1025)}} & \makecell{90.2 \\ \textcolor{gray!80}{(7042/7805)}} \\
& \textbf{Splatfacto-i (Ours)}  & \makecell{\textbf{98.2} \\ \textcolor{gray!80}{(\textbf{6372}/6492)}} & \makecell{\textbf{97.0} \\ \textcolor{gray!80}{(\textbf{6394}/6589)}} & \makecell{\textbf{89.2} \\ \textcolor{gray!80}{(\textbf{7226}/8101)}} & \makecell{\textbf{91.3} \\ \textcolor{gray!80}{(\textbf{6984}/7648)}} \\
\bottomrule
\end{tabular}
\begin{tablenotes}[flushleft]\footnotesize
\item [*] : Partial map due to performance limitations.
\end{tablenotes}
\end{threeparttable}
\end{table}

We assumed that in the normal map, surfaces with predominantly upward-facing normals would indicate traversable space. However, as illustrated in Fig.~\ref{fig:recon_quantitative}, a foundation model struggle to produce clean surface normals for the ground due to artifacts and noise. Therefore, as evidenced by Table~\ref{tab:valid_node_edge_ratios}, our model produces a more valid map.

We assumed that in the depth map, regions without obstacles would correspond to traversable space. However, as shown in Fig.~\ref{fig:recon_quantitative}, artifacts in the representation were often misinterpreted as obstacles, leading to false negatives. Therefore, we did not actively rely on the depth map.

As shown in Table~\ref{tab:valid_node_edge_ratios}, our Splatfacto-i model achieves the best performance when utilizing both the alpha and normal maps. Using either map alone resulted in several failure cases. For instance, the alpha map was unreliable in areas where reconstruction failed to generate Gaussians, such as on texture-less surfaces or incorrectly masked floors (Fig.~\ref{fig:why_ensamble_is_important}(a)). Conversely, the normal map failed on a full-height window, where the foundation model extended the floor plane across the glass, leading to incorrect surface normals. (Fig.~\ref{fig:why_ensamble_is_important}(b)). Therefore, we found that ensembling both alpha and normal maps is the most robust method for generating a valid map.

\subsection{R2R Task in NVSim Environments}
\begin{table}[h!]
\caption{R2R Navigation Task Performance Comparison on NVSim (COEX Dataset - Val unseen)}
\label{tab:my_table}
\scriptsize 
\centering 
\setlength{\tabcolsep}{2pt} 
\renewcommand{\arraystretch}{1} 
\begin{tabular}{l|ccccc}
\toprule
\textbf{Methods} & \textbf{NE $\downarrow$} & \textbf{SR(\%)$\uparrow$} & \textbf{OSR(\%)$\uparrow$} & \textbf{SPL(\%)$\uparrow$} & \textbf{PL} \\
\midrule
SHORTEST & 0.0 & 100 & 100 & 100 & 19.02 \\
RANDOM & 16.429 & 0.997 & 2.1 & 0.683 & 27.087 \\
\midrule
VLN$\circlearrowright$BERT\cite{hong2021vln}   & 10.002 & 17.84 & 23.75 & 17.26 & 16.2879 \\
VLN$\circlearrowright$BERT \texttt{(val\_seen)} & 7.705 & 22.77 & 34.27 & 21.98 & 16.473 \\
\midrule
MapGPT (with GPT4o)\cite{chen2024mapgpt}  & 10.77 & 12.5 & 27.78 & 10.97 & 20.72 \\
MapGPT (with GPT5)\cite{chen2024mapgpt} & 9.92 & 17.5 & 30.5 & 14.37 & 23.348 \\
\bottomrule
\end{tabular}
\end{table}
\begin{figure}[t]
    \centering
    \includegraphics[width=0.7\linewidth]{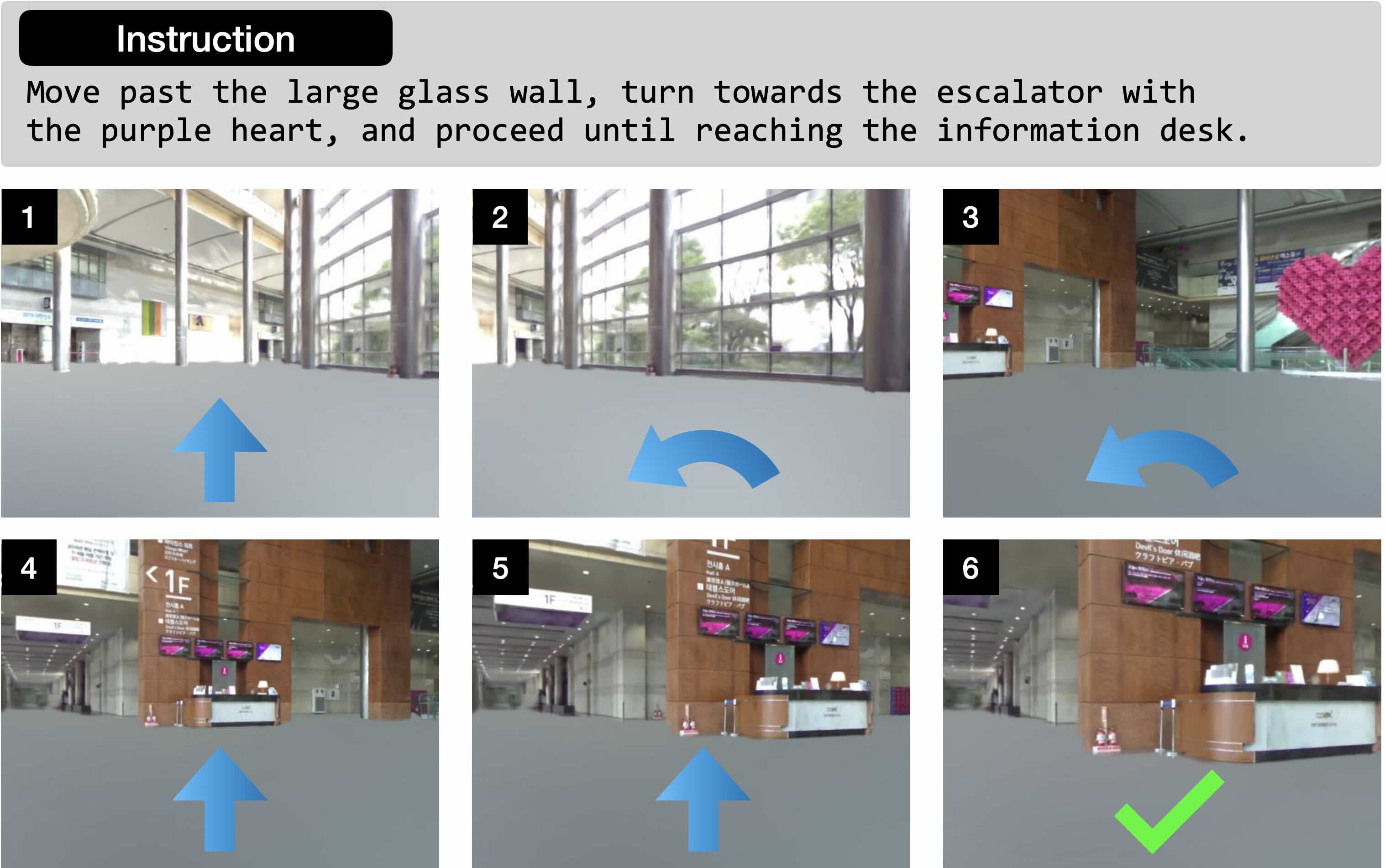}
    \caption{Example from our R2R dataset showing an image sequence of a navigation path with its natural language instruction}
    \label{fig:instr_fig}
\end{figure}
In this experiment, we perform established Room-to-Room (R2R) tasks within our reconstructed environment. Unlike previous experiments which were divided into 15 submaps, we reconfigured the environment into 10 submaps for this study. This was done to verify the feasibility of navigation between submaps in a practical R2R task.
To adapt the VLN R2R task for our dataset, instructions for each path were required. For this purpose, we utilized the GPT-4o model, which is capable of processing both images and text, to generate these instructions (Fig.~\ref{fig:instr_fig}).
The evaluation metrics employed include Path Length (PL), Navigation Error (NE), Success Rate (SR), Oracle Success Rate (OSR), and Success Rate Penalized by Path Length (SPL).
For the R2R experiment, we utilized VLN$\circlearrowright$BERT\cite{hong2021vln}, which requires training, and MapGPT\cite{chen2024mapgpt}, which allows for zero-shot validation. We also compare against standard learning-free baselines: RANDOM, which takes 10 random actions, and SHORTEST, which follows the optimal path.
Table~\ref{tab:my_table} demonstrates that the conventional R2R task can be performed in our environment. As shown in the table~\ref{tab:my_table}, MapGPT tends to explore continuously when the description is insufficient (20.72m with MapGPT and 16.28m with VLN Bert), resulting in a lower SPL performance (10.97\% with MapGPT and 17.26\% with VLN Bert). Furthermore, the learning-free baselines exhibit weaker performance than on the standard R2R dataset (16.3\% SR with the R2R dataset and 0.9\% SR with our dataset) \cite{anderson2018vision}.

These results indicate that the task in our large-scale environment is more challenging than those conducted on conventional indoor benchmarks. The performance gap in VLN$\circlearrowright$BERT between the \texttt{val\_seen} and \texttt{val\_unseen} splits is due to the characteristics of our dataset.
Our dataset, covering a large-scale environment, is partitioned via distance-based clustering, which creates a high degree of visual and structural disparity between seen and unseen submaps. Unlike the homogeneous environments in standard R2R datasets, our method presents a greater generalization challenge.

\section{Conclusion}
In summary, we proposed NVSim, a novel framework that automatically constructs large-scale, navigable indoor simulators directly from image sequences. Our method creates robustly navigable simulators without costly 3D scans or scaling issues by ensuring clean floor representations and automatically discovering traversable routes. We validate this system's capability by generating large-scale, accurate navigation graphs from real-world traversals without a mesh.

\bibliographystyle{IEEEtranS}
\bibliography{citation}

\begin{thebibliography}{10}
\providecommand{\url}[1]{#1}
\csname url@rmstyle\endcsname
\providecommand{\newblock}{\relax}
\providecommand{\bibinfo}[2]{#2}
\providecommand\BIBentrySTDinterwordspacing{\spaceskip=0pt\relax}
\providecommand\BIBentryALTinterwordstretchfactor{4}
\providecommand\BIBentryALTinterwordspacing{\spaceskip=\fontdimen2\font plus
\BIBentryALTinterwordstretchfactor\fontdimen3\font minus \fontdimen4\font\relax}
\providecommand\BIBforeignlanguage[2]{{%
\expandafter\ifx\csname l@#1\endcsname\relax
\typeout{** WARNING: IEEEtran.bst: No hyphenation pattern has been}%
\typeout{** loaded for the language `#1'. Using the pattern for}%
\typeout{** the default language instead.}%
\else
\language=\csname l@#1\endcsname
\fi
#2}}

\bibitem{adamkiewicz2022vision}
M.~Adamkiewicz, T.~Chen, A.~Caccavale, R.~Gardner, P.~Culbertson, J.~Bohg, and M.~Schwager, ``Vision-only robot navigation in a neural radiance world,'' \emph{IEEE Robotics and Automation Letters}, vol.~7, no.~2, pp. 4606--4613, 2022.

\bibitem{anderson2018vision}
P.~Anderson, Q.~Wu, D.~Teney, J.~Bruce, M.~Johnson, N.~S{\"u}nderhauf, I.~Reid, S.~Gould, and A.~Van Den~Hengel, ``Vision-and-language navigation: Interpreting visually-grounded navigation instructions in real environments,'' in \emph{Proceedings of the IEEE conference on computer vision and pattern recognition}, 2018, pp. 3674--3683.

\bibitem{chang2017matterport3d}
A.~Chang, A.~Dai, T.~Funkhouser, M.~Halber, M.~Niessner, M.~Savva, S.~Song, A.~Zeng, and Y.~Zhang, ``Matterport3d: Learning from rgb-d data in indoor environments,'' \emph{arXiv preprint arXiv:1709.06158}, 2017.

\bibitem{chen2024mapgpt}
J.~Chen, B.~Lin, R.~Xu, Z.~Chai, X.~Liang, and K.-Y.~K. Wong, ``Mapgpt: Map-guided prompting with adaptive path planning for vision-and-language navigation,'' \emph{arXiv preprint arXiv:2401.07314}, 2024.

\bibitem{chen2025splat}
T.~Chen, O.~Shorinwa, J.~Bruno, A.~Swann, J.~Yu, W.~Zeng, K.~Nagami, P.~Dames, and M.~Schwager, ``Splat-nav: Safe real-time robot navigation in gaussian splatting maps,'' \emph{IEEE Transactions on Robotics}, 2025.

\bibitem{gu2024survey}
J.~Gu, X.~Jiang, Z.~Shi, H.~Tan, X.~Zhai, C.~Xu, W.~Li, Y.~Shen, S.~Ma, H.~Liu, \emph{et~al.}, ``A survey on llm-as-a-judge,'' \emph{arXiv preprint arXiv:2411.15594}, 2024.

\bibitem{hong2021vln}
Y.~Hong, Q.~Wu, Y.~Qi, C.~Rodriguez-Opazo, and S.~Gould, ``Vln bert: A recurrent vision-and-language bert for navigation,'' in \emph{Proceedings of the IEEE/CVF conference on Computer Vision and Pattern Recognition}, 2021, pp. 1643--1653.

\bibitem{hornung2013octomap}
A.~Hornung, K.~M. Wurm, M.~Bennewitz, C.~Stachniss, and W.~Burgard, ``Octomap: An efficient probabilistic 3d mapping framework based on octrees,'' \emph{Autonomous robots}, vol.~34, no.~3, pp. 189--206, 2013.

\bibitem{kerbl2023gaussiansplatting}
\BIBentryALTinterwordspacing
B.~Kerbl, G.~Kopanas, T.~Leimkuehler, and G.~Drettakis, ``3d gaussian splatting for real-time radiance field rendering,'' \emph{ACM Trans. Graph.}, vol.~42, no.~4, July 2023. [Online]. Available: \url{https://doi.org/10.1145/3592433}
\BIBentrySTDinterwordspacing

\bibitem{ku2020room}
A.~Ku, P.~Anderson, R.~Patel, E.~Ie, and J.~Baldridge, ``Room-across-room: Multilingual vision-and-language navigation with dense spatiotemporal grounding,'' \emph{arXiv preprint arXiv:2010.07954}, 2020.

\bibitem{lee2021large}
D.~Lee, S.~Ryu, S.~Yeon, Y.~Lee, D.~Kim, C.~Han, Y.~Cabon, P.~Weinzaepfel, N.~Gu{\'e}rin, G.~Csurka, \emph{et~al.}, ``Large-scale localization datasets in crowded indoor spaces,'' in \emph{Proceedings of the IEEE/CVF Conference on Computer Vision and Pattern Recognition}, 2021, pp. 3227--3236.

\bibitem{mildenhall2021nerf}
B.~Mildenhall, P.~P. Srinivasan, M.~Tancik, J.~T. Barron, R.~Ramamoorthi, and R.~Ng, ``Nerf: Representing scenes as neural radiance fields for view synthesis,'' \emph{Communications of the ACM}, vol.~65, no.~1, pp. 99--106, 2021.

\bibitem{qiao2025open}
Y.~Qiao, W.~Lyu, H.~Wang, Z.~Wang, Z.~Li, Y.~Zhang, M.~Tan, and Q.~Wu, ``Open-nav: Exploring zero-shot vision-and-language navigation in continuous environment with open-source llms,'' in \emph{2025 IEEE International Conference on Robotics and Automation (ICRA)}.\hskip 1em plus 0.5em minus 0.4em\relax IEEE, 2025, pp. 6710--6717.

\bibitem{ramakrishnan2021habitat}
S.~K. Ramakrishnan, A.~Gokaslan, E.~Wijmans, O.~Maksymets, A.~Clegg, J.~Turner, E.~Undersander, W.~Galuba, A.~Westbury, A.~X. Chang, \emph{et~al.}, ``Habitat-matterport 3d dataset (hm3d): 1000 large-scale 3d environments for embodied ai,'' \emph{arXiv preprint arXiv:2109.08238}, 2021.

\bibitem{saura2021spherical}
M.~Saura-Herreros, A.~Lopez, and J.~Ribelles, ``Spherical panorama compositing through depth estimation,'' \emph{The Visual Computer}, vol.~37, no.~9, pp. 2809--2821, 2021.

\bibitem{savva2019habitat}
M.~Savva, A.~Kadian, O.~Maksymets, Y.~Zhao, E.~Wijmans, B.~Jain, J.~Straub, J.~Liu, V.~Koltun, J.~Malik, \emph{et~al.}, ``Habitat: A platform for embodied ai research,'' in \emph{Proceedings of the IEEE/CVF international conference on computer vision}, 2019, pp. 9339--9347.

\bibitem{shen2021igibson}
B.~Shen, F.~Xia, C.~Li, R.~Mart{\'\i}n-Mart{\'\i}n, L.~Fan, G.~Wang, C.~P{\'e}rez-D’Arpino, S.~Buch, S.~Srivastava, L.~Tchapmi, \emph{et~al.}, ``igibson 1.0: A simulation environment for interactive tasks in large realistic scenes,'' in \emph{2021 IEEE/RSJ International Conference on Intelligent Robots and Systems (IROS)}.\hskip 1em plus 0.5em minus 0.4em\relax IEEE, 2021, pp. 7520--7527.

\bibitem{tancik2022block}
M.~Tancik, V.~Casser, X.~Yan, S.~Pradhan, B.~Mildenhall, P.~P. Srinivasan, J.~T. Barron, and H.~Kretzschmar, ``Block-nerf: Scalable large scene neural view synthesis,'' in \emph{Proceedings of the IEEE/CVF conference on computer vision and pattern recognition}, 2022, pp. 8248--8258.

\bibitem{tao2024silvr}
Y.~Tao, Y.~Bhalgat, L.~F.~T. Fu, M.~Mattamala, N.~Chebrolu, and M.~Fallon, ``Silvr: Scalable lidar-visual reconstruction with neural radiance fields for robotic inspection,'' in \emph{2024 IEEE International Conference on Robotics and Automation (ICRA)}.\hskip 1em plus 0.5em minus 0.4em\relax IEEE, 2024, pp. 17\,983--17\,989.

\bibitem{wu2025sparse2dgs}
J.~Wu, R.~Li, Y.~Zhu, R.~Guo, J.~Sun, and Y.~Zhang, ``Sparse2dgs: Geometry-prioritized gaussian splatting for surface reconstruction from sparse views,'' in \emph{Proceedings of the Computer Vision and Pattern Recognition Conference}, 2025, pp. 11\,307--11\,316.

\bibitem{xu2024splatfacto}
C.~Xu, J.~Kerr, and A.~Kanazawa, ``Splatfacto-w: A nerfstudio implementation of gaussian splatting for unconstrained photo collections,'' \emph{arXiv preprint arXiv:2407.12306}, 2024.

\bibitem{yeshwanth2023scannetpp}
C.~Yeshwanth, Y.-C. Liu, M.~Nie{\ss}ner, and A.~Dai, ``Scannet++: A high-fidelity dataset of 3d indoor scenes,'' in \emph{Proceedings of the IEEE/CVF International Conference on Computer Vision}, 2023, pp. 12--22.

\bibitem{younis2025sparse}
T.~Younis and Z.~Cheng, ``Sparse-view 3d reconstruction: Recent advances and open challenges,'' \emph{arXiv preprint arXiv:2507.16406}, 2025.

\bibitem{zhou2024navgpt}
G.~Zhou, Y.~Hong, and Q.~Wu, ``Navgpt: Explicit reasoning in vision-and-language navigation with large language models,'' in \emph{Proceedings of the AAAI Conference on Artificial Intelligence}, vol.~38, no.~7, 2024, pp. 7641--7649.

\bibitem{zhu2021soon}
F.~Zhu, X.~Liang, Y.~Zhu, Q.~Yu, X.~Chang, and X.~Liang, ``Soon: Scenario oriented object navigation with graph-based exploration,'' in \emph{Proceedings of the IEEE/CVF Conference on Computer Vision and Pattern Recognition}, 2021, pp. 12\,689--12\,699.

\bibitem{zhu2025vr}
S.~Zhu, L.~Mou, D.~Li, B.~Ye, R.~Huang, and H.~Zhao, ``Vr-robo: A real-to-sim-to-real framework for visual robot navigation and locomotion,'' \emph{IEEE Robotics and Automation Letters}, 2025.

\end{thebibliography}
\end{document}